\documentclass[letterpaper]{article} 
\usepackage{aaai24}  
\usepackage{times}  
\usepackage{helvet}  
\usepackage{courier}  
\usepackage[hyphens]{url}  
\usepackage{graphicx} 
\usepackage{natbib}  
\usepackage{caption} 
\usepackage{subcaption}
\usepackage{algorithm}
\usepackage{algorithmic}
\usepackage{amsmath,amssymb}
\usepackage{multirow}
\usepackage{adjustbox}
\usepackage{booktabs}
\usepackage{newfloat}
\usepackage{listings}
\DeclareCaptionStyle{ruled}{labelfont=normalfont,labelsep=colon,strut=off} 
\floatstyle{ruled}
\newfloat{listing}{tb}{lst}{}
\floatname{listing}{Listing}
\title{SSLCL: An Efficient Model-Agnostic Supervised Contrastive Learning Framework for Emotion Recognition in Conversations}
\author{
    Tao Shi\textsuperscript{\rm 1}\equalcontrib,
    Xiao Liang\textsuperscript{\rm 1}\equalcontrib,
    Yaoyuan Liang\textsuperscript{\rm 1},
    Xinyi Tong\textsuperscript{\rm 1},
    Shao-Lun Huang\textsuperscript{\rm 1}\thanks{\; Corresponding author.}
}
\affiliations{
    \textsuperscript{\rm 1}Tsinghua Shenzhen International Graduate School, Tsinghua University\\
    \{shitao21,  liangx22, liang-yy21, txy18\}@mails.tsinghua.edu.cn, 
    shaolun.huang@sz.tsinghua.edu.cn

}

\usepackage{bibentry}

\begin{document}

\maketitle

\begin{abstract}
Emotion recognition in conversations (ERC) is a rapidly evolving task within the natural language processing community, which aims to detect the emotions expressed by speakers during a conversation. Recently, a growing number of ERC methods have focused on leveraging supervised contrastive learning (SCL) to enhance the robustness and generalizability of learned features. However, current SCL-based approaches in ERC are impeded by the constraint of large batch sizes and the lack of compatibility with most existing ERC models. To address these challenges, we propose an efficient and model-agnostic SCL framework named Supervised Sample-Label Contrastive Learning with Soft-HGR Maximal Correlation (SSLCL), which eliminates the need for a large batch size and can be seamlessly integrated with existing ERC models without introducing any model-specific assumptions. Specifically, we introduce a novel perspective on utilizing label representations by projecting discrete labels into dense embeddings through a shallow multilayer perceptron, and formulate the training objective to maximize the similarity between sample features and their corresponding ground-truth label embeddings, while minimizing the similarity between sample features and label embeddings of disparate classes. Moreover, we innovatively adopt the Soft-HGR maximal correlation as a measure of similarity between sample features and label embeddings, leading to significant performance improvements over conventional similarity measures. Additionally, multimodal cues of utterances are effectively leveraged by SSLCL as data augmentations to boost model performances. Extensive experiments on two ERC benchmark datasets, IEMOCAP and MELD, demonstrate the compatibility and superiority of our proposed SSLCL framework compared to existing state-of-the-art SCL methods. Our code is available at \url{https://github.com/TaoShi1998/SSLCL}.
\end{abstract}

\section{Introduction}
Emotion recognition in conversations (ERC) aims to accurately identify the emotion conveyed in each utterance during a conversation. The widespread potential applications in areas such as opinion mining \cite{majumder2019dialoguernn}, empathetic chatbots \cite{zhang-etal-2023-dualgats}, and medical diagnosis \cite{hu-etal-2021-mmgcn} make ERC an essential task within the field of natural language processing.

To address the problem of ERC,  an increasing number of studies \cite{9706271, 9938005, li2022contrast, song-etal-2022-supervised, shi-huang-2023-multiemo, yang-etal-2023-self, yang2023cluster, tu-etal-2023-context} have leveraged supervised contrastive learning (SCL) \cite{khosla2020supervised} to learn robust and generalized feature representations that are better suited for emotion classifications. Supervised contrastive learning is a representation learning technique that aims to enhance the robustness of learned representations through pulling samples belonging to the same category (positive pairs) closer, while simultaneously pushing apart samples from disparate classes (negative pairs). Currently, SCL-based approaches in ERC can be broadly classified into two categories: (1) Vanilla supervised contrastive loss (SupCon loss) \cite{9706271, 9938005, yang-etal-2023-self}; (2) Variants of the SupCon loss that integrate SCL with other learning strategies, including prototypical networks and curriculum learning \cite{song-etal-2022-supervised}, multiview representation learning \cite{li2022contrast}, hard example mining \cite{shi-huang-2023-multiemo}, self-supervised contrastive learning \cite{tu-etal-2023-context}, and emotion prototypes \cite{yang2023cluster}. 

\begin{figure}
\centering
\includegraphics[width=0.48\textwidth]{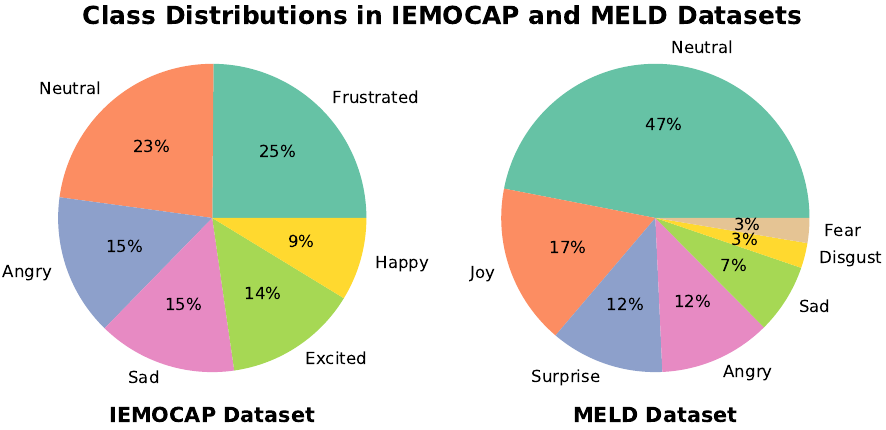}
\caption{Illustration of imbalanced class distributions in IEMOCAP and MELD.}\label{fig:1}
\end{figure}

Despite the surging popularity of applying SCL to ERC, several challenges still remain unresolved: (1) \textbf{ERC models using the vanilla SupCon loss typically require a large batch size to achieve satisfactory performances, which is computationally expensive.} Existing widely-used ERC benchmark datasets, such as IEMOCAP \cite{busso2008iemocap} and MELD \cite{poria-etal-2019-meld}, are plagued by the class imbalance problem. As illustrated in Figure \ref{fig:1}, Both IEMOCAP and MELD are class-imbalanced, with MELD being particularly skewed towards certain classes. Thus, when utilizing SupCon on these datasets, a large batch size is often needed to guarantee that each training sample from minority classes has at least one positive pair within the batch, otherwise the loss can not be properly calculated, leading to a significant degradation in model performance \cite{shi-huang-2023-multiemo}. (2) \textbf{Variants of the SupCon loss tend to have difficulty integrating with most existing ERC approaches.} While combining SCL with other learning techniques allows many SupCon variants to overcome the constraint of a large batch size, these SupCon variants tend to be incompatible with the majority of existing ERC frameworks, since they depend on specific modeling assumptions \cite{song-etal-2022-supervised, tu-etal-2023-context, yang2023cluster} that can not be easily extended to other models. Moreover, SupCon variants often suffer from an excessive level of complexity. (3) \textbf{The inherent multimodal nature of utterances in ERC has not been leveraged by existing SCL-based algorithms.} A distinctive characteristic of the ERC task is that each utterance is accompanied by its corresponding textual, audio and visual modalities. However, the potential of leveraging multimodal cues has remained unexplored by existing SCL literature.

To tackle the aforementioned challenges, in this paper, we propose a novel SCL framework named \textbf{Supervised Sample-Label Contrastive Learning with Soft-HGR Maximal Correlation (SSLCL)}, which is an efficient and model-agnostic approach that eliminates the need for a large batch size and can be seamlessly integrated with existing ERC models without introducing any model-specific assumptions. Additionally, multimodal information is effectively leveraged by SSLCL as data augmentation to achieve more promising performances.

Specifically, the key insight of our work is a fundamental shift in how labels are utilized: in contrast to the prevalent use of one-hot vector label representations in existing SCL-based approaches, we embed each discrete emotion category into a dense label embedding through a shallow multilayer perceptron, and utilize the learned label embedding of a specific class as the ``ground-truth representation" for samples belonging to that class. Then, we formulate the training objective to maximize the similarity between sample features and their corresponding ground-truth label embeddings (positive sample-label pairs), while minimizing the similarity between sample features and label embeddings of different classes (negative sample-label pairs). Furthermore, inspired by Cutout \cite{devries2017improved}, a commonly-used image augmentation technique in the computer vision community, we utilize multimodal views of the same sample as its data augmentations. Consequently, each training sample is guaranteed to have multiple positive sample-label pairs within the batch, regardless of the batch size. \cite{lee-lee-2022-compm}

Another notable contribution of SSLCL that distinguishes it from existing SCL methods is the adaptation of the soft-Hirschfeld-Gebelein-R{\'e}nyi (Soft-HGR) maximal correlation \cite{wang2019efficient} as a measure of similarity between sample features and label embeddings. Soft-HGR aims to extract maximally correlated representations across multiple random variables. In our SSLCL setting, we utilize Soft-HGR as a similarity measure to capture the complex correlations between sample-label pairs, leading to significant performance improvements over conventional similarity measures (refer to the Results and Analysis section). 

Finally, to demonstrate the compatibility and superiority of SSLCL, we conduct extensive experiments on two ERC benchmark datasets: IEMOCAP and MELD. Experimental results provide compelling evidence that SSLCL exhibits three key advantages against existing SCL literature: (1) SSLCL can be easily integrated with existing ERC models using any methodologies, without introducing any model-specific assumptions; (2) ERC methods using SSLCL can achieve excellent performances without the need for a large batch size; (3) The utilization of SSLCL yields significant improvements in model performances, surpassing existing SCL-based approaches and achieving new state-of-the-art results on both IEMOCAP and MELD. 

To summarize, the main contributions of this work are four-fold: (1) Through a novel utilization of label representations, we effectively address the constraints posed by large batch sizes and incompatibility with most existing ERC architectures encountered in current SCL-based methods; (2) To the best of our knowledge, we are the first in the SCL community to leverage Soft-HGR as a measure of similarity, which yields considerable performance improvements over traditional similarity measures; (3) We innovatively leverage multimodal information as data augmentation to enhance model performances; (4) Extensive experiments on two ERC benchmark datasets, IEMOCAP and MELD, demonstrate the compatibility and superiority of SSLCL compared with existing state-of-the-art SCL approaches.

\section{Related Work}

\subsection{Supervised Contrastive Learning}
Supervised contrastive learning \cite{khosla2020supervised} is a representation learning technique that extends self-supervised contrastive learning (self-supervised CL) \cite{wu2018unsupervised} by leveraging label information, in which samples from the same class are pulled closer while samples belonging to different classes are pushed apart within the batch. 

Recently, an increasing number of ERC models have utilized the vanilla SupCon loss to enhance model performances. For example, Sentic GAT \cite{9706271} leverages SupCon to discriminate context-free and context-sensitive representations. \citeauthor{9938005} (2022) employs SupCon to enhance cross-modal consistency. SCMM \cite{yang-etal-2023-self} utilizes the SupCon loss to improve the discriminability of learned multimodal features. However, vanilla SupCon in ERC typically requires a large batch size to achieve satisfactory performances, rendering it computationally expensive. 

 
\subsection{Emotion Recognition in Conversations}
\subsubsection{Recurrence-based Methods} 
BC-LSTM \cite{poria-etal-2017-context} captures conversational contexts based on long short-term memory (LSTM) networks. DialogueRNN \cite{majumder2019dialoguernn} employs three gated recurrent units (GRUs) to model the context and keep track of speaker states. Inspired by the cognitive theory of emotion, DialogueCRN \cite{hu-etal-2021-dialoguecrn} introduces a cognitive perspective to comprehend contextual information based on LSTMs. 

\subsubsection{Transformer-based Methods} 
HiTrans \cite{li-etal-2020-hitrans} utilizes two hierarchical transformers to capture local- and global-level utterance representations. EmoBERTa \cite{kim2021emoberta} learns inter- and intra-speaker states and conversational contexts through fine-tuning a pre-trained RoBERTa. CoG-BART is proposed by \citeauthor{li2022contrast} (2022), which leverages a pretrained encoder-decoder model BART to capture contextual information. CoMPM \cite{lee-lee-2022-compm} utilizes a transformer-encoder based context model and a speaker-aware pretrained memory module to capture conversational contexts. 


\subsubsection{Graph-based Methods} 
DialogueGCN \cite{ghosal-etal-2019-dialoguegcn} models conversations based on a graph convolutional network (GCN). DAG-ERC \cite{shen-etal-2021-directed} encodes utterances through a directed acyclic graph (DAG) to better understand the inherent structure within a conversation. M3Net \cite{chen2023multivariate} leverages graph neural networks (GNNs) to capture multivariate relationships among multiple modalities and conversational contexts. 

\subsubsection{SCL-based Methods} 
The increasing popularity of SCL has led to numerous attempts to combine SCL with other learning strategies. To illustrate, SPCL \cite{song-etal-2022-supervised} integrates SCL with prototypical networks and curriculum learning to alleviate the imbalanced class problem without requiring a large batch size. However, SPCL is incompatible with most existing ERC models because its difficulty measure function requires the emotion representations learned by the ERC model to be distance-aware. CoG-BART \cite{li2022contrast} introduces multiview-SupCon to addresses the batch size constraint associated with SupCon, in which each sample is paired with a gradient-detached copy.  Nevertheless, a significant drawback of multiview-SupCon is that it calculates the similarity between identical samples, which can lead to suboptimal representations and potentially hinder the model's ability to generalize effectively. MultiEMO \cite{shi-huang-2023-multiemo} introduces a sample-weighted focal contrastive (SWFC) loss, which combines SCL with hard sample mining to alleviate the difficulty of recognizing minority and semantically similar emotions. However, similar to vanilla SCL, the SWFC loss also relies on a significantly large batch size to achieve desirable results. CKCL \cite{tu-etal-2023-context} integrates SCL with self-supervised CL to differentiate context- and knowledge-independent utterances. Nonetheless, a major limitation of CKCL is it can not be applied to most ERC models that do not rely on external knowledge. SCCL \cite{yang2023cluster} combines SCL with valence-arousal-dominance (VAD) emotion Prototypes to boost model performances. However, SCCL only considers context-independent word-level VAD, whereas the majority of existing ERC approaches are context-dependent.

\begin{figure*}
\centering
\includegraphics[width=0.95\textwidth]{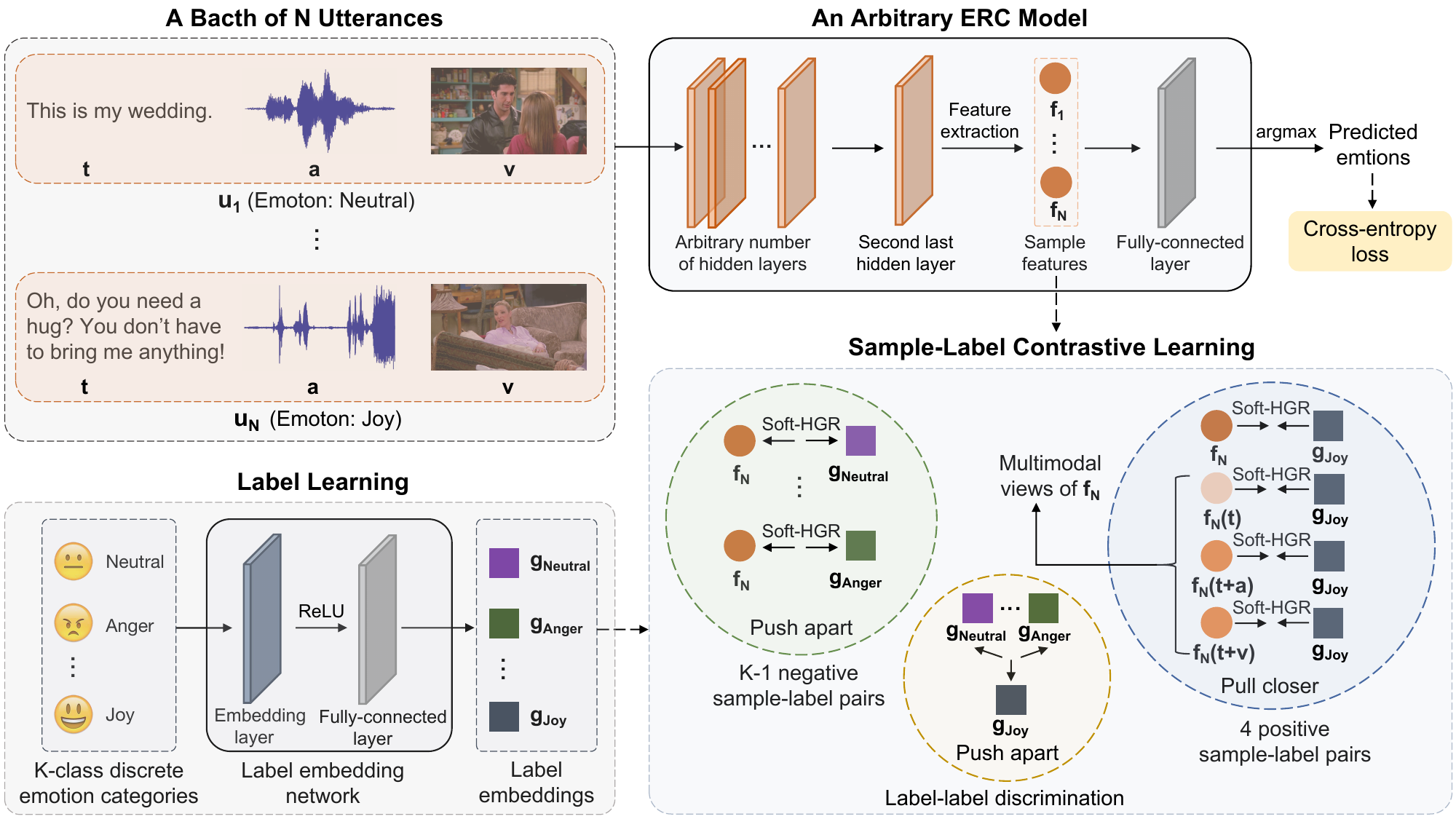}
\caption{Illustration of the overall framework of SSLCL\protect\footnotemark, which consists of three key components: sample feature extraction, label learning, and sample-label contrastive learning. For the sake of simplicity, sample-label contrastive learning is illustrated using the $N$-th utterance in the batch. In this Figure, $t$, $a$ and $v$ represent the textual, audio and visual modalities of utterances respectively, $\textbf{f}_{N}(t), \textbf{f}_{N}(t + a), \textbf{f}_{N}(t + v)$ are data augmentations of the $N$-th sample feature with different missing modalities.}\label{fig:2}
\end{figure*}
\section{Preliminary}
\subsection{Soft-HGR Maximum Correlation}
Soft-Hirschfeld-Gebelein-R{\'e}nyi (Soft-HGR) maximum correlation \cite{wang2019efficient} is an extension of the HGR maximum correlation \cite{hirschfeld1935connection, gebelein1941statistische, renyi1959measures}, which is a measure of dependence between random variables. Soft-HGR preserves the same feature geometry as in the HGR maximal correlation, while simultaneously avoiding strict whitening constraints imposed by HGR. 

Soft-HGR aims to extract maximally correlated feature representations from multiple random modalities. Formally, given two random variables represented by $X$ and $Y$, the objective of Soft-HGR is to learn nonlinear feature mappings $\textbf{f}(X)$ and $\textbf{g}(Y)$ that are maximally correlated. The Soft-HGR objective is defined as follows:
\begin{equation}
\begin{aligned}
\max_{\textbf{f}, \textbf{g}} \quad & \mathbb{E}[\textbf{f}^\mathrm{T}(X)\textbf{g}(Y)] - \frac{1}{2}\text{tr}(\text{cov}(\textbf{f}(X))\text{cov}(\textbf{g}(Y))), \label{eq:1}\\
\textrm{s.t.} \quad & \mathbb{E}[\textbf{f}(X)] = \mathbb{E}[\textbf{g}(Y)] = 0.
\end{aligned}  
\end{equation}

As illustrated in equation \ref{eq:1}, Soft-HGR consists of two inner products: one between feature mappings and another between feature covariances. The first inner product is consistent with the objective of the HGR maximal correlation, while the second one serves as a soft regularizer that replaces the hard whitening constraints in HGR.

From an information theory perspective, the optimal nonlinear feature transformations $\textbf{f}(X)$ and $\textbf{g}(Y)$ learned through Soft-HGR capture the maximum amount of mutual information shared by random variables $X$ and $Y$, rendering them highly informative and well-suited for downstream discriminative tasks.

\section{Methodology}

\subsection{Problem Definition}
Given a conversation which consists of $n$ utterances $\textbf{u}_{1}, \textbf{u}_{2}, \dots, \textbf{u}_{n}$ expressed by speakers $S_{\textbf{u}_{1}}, S_{\textbf{u}_{2}}, \dots, S_{\textbf{u}_{n}}$, the objective of ERC is to assign an emotion label from a predefined $K$-class emotion category set $\mathcal{Y}$ to each utterance in this conversation. Each utterance is accompanied by its corresponding textual ($t$), audio ($a$) and visual ($v$) modalities, which can be represented as follows:
\begin{equation}
\begin{aligned}
\textbf{u}_{i} = \{\textbf{u}_{i}^{t}, \textbf{u}_{i}^{a}, \textbf{u}_{i}^{v}\}, i \in \{1, \dots, n\},
\end{aligned}  
\label{eq:2}
\end{equation}
where $\textbf{u}_{i}^{t} \in \mathbb{R}^{d_{t}}, \textbf{u}_{i}^{a} \in \mathbb{R}^{d_{a}}, \textbf{u}_{i}^{v} \in \mathbb{R}^{d_{v}}$.

\subsection{Overview of the Proposed SSLCL Framework}
The overall framework of SSLCL is illustrated in Figure \ref{fig:2}, which is made up of three key components: sample feature extraction, label learning, and sample-label contrastive learning. We discuss each component in detail as follows.

\subsection{Sample Feature Extraction}
A major advantage of SSLCL over established SCL literature is its ability to seamlessly integrate with existing ERC approaches without introducing any model-specific assumptions. As shown in Figure \ref{fig:2}, this is achieved through a simple yet effective approach: for any ERC model, we extract the sample features from its second-last hidden layer and contrast them with label embeddings learned through a label embedding network, which is jointly optimized with the ERC model during training. There are no architectural modifications or modeling assumptions required in this process.

\subsection{Label Learning} 
In contrast to the commonly adopted one-hot vector label representations in existing SCL-based approaches, our proposed SSLCL framework innovatively projects each discrete emotion category into a dense embedding through a simple yet powerful two-layer multilayer perceptron (MLP), which consists of a embedding layer, a fully-connected layer and a rectified linear unit (ReLU). Although a deeper and more complex label embedding network \cite{sun2017label} could be adopted, experimental results demonstrate that a two-layer shallow MLP is sufficient to achieve satisfactory performance (refer to Appendix D). 
\footnotetext{Our SSLCL framework can be applied to ERC models with any modalities. We use the multimodal case in Figure \ref{fig:2} in order to illustrate how data augmentation works.}

Specifically, given a $K$-class discrete emotion category set $\mathcal{Y} = \{1, 2, \dots, K\}$, the label embedding $\textbf{g}_{i} \in \mathbb{R}^{d}$ for the $i$-th emotion label is obtained as follows:
\begin{align}
& \textbf{e}_{i} = \max(0, \text{EmbeddingLayer}(i)), \\
& \textbf{g}_{i} = \textbf{W}_{g}\textbf{e}_{i} + \textbf{b}_{g},
\end{align}  
where $\textbf{e}_{i} \in \mathbb{R}^{d_{e}}$ is the hidden output, $\textbf{W}_{g} \in \mathbb{R}^{d \times d_{e}}$ is the weight matrix, and $\textbf{b}_{g} \in \mathbb{R}^{d}$ is the bias parameter. 

After projecting emotion labels into label embeddings, we utilize the label embedding $\textbf{g}_{i}$ as the “ground-truth representation” for samples belonging to class $i$, and formulate the training objective to maximize the similarity between sample features and their corresponding ground-truth label embeddings, while simultaneously minimizing the similarity between sample features and label embeddings of different classes. This novel formulation guarantees that each training sample is accompanied by at least one positive feature-label pair within each batch, regardless of the batch size.

\begin{table*}[t]
    \centering 
    \large  
    \resizebox{1\linewidth}{!}{
    \begin{tabular}{l|ccccccc|cccccccc}
      \toprule
      \multirow{2}{*}{Methods} & \multicolumn{7}{c|}{IEMOCAP} & \multicolumn{8}{c}{MELD} \\ 
      & Happy & Sad & Neutral & Angry & Excited & Frustrated & \textbf{w-F1} & Neutral & Surprise & Fear & Sad & Joy & Disgust & Angry & \textbf{w-F1} \\
      \midrule
        DialogueRNN & 33.67 & 72.91 & 52.32 & 61.40 & 74.24 & 56.54 & 59.75 & 75.50 & 48.81 & 0.00 & 18.24 & 52.04 & 0.00 & 45.77 & 57.11 \\ 
        + SupCon & 34.15 & 70.48 & 54.50 & 62.31 & 64.77 & 60.83 & 59.30 & 75.77 & 46.56 & 0.00 & 20.29 & 52.76 & 0.00 & 45.91 & 57.29 \\ 
        + mv-SupCon & 30.43 & 68.33 & 52.49 & 59.57 & 69.55 & 61.37 & 58.89 & 75.80 & 47.62 & 0.00 & 18.24 & 51.96 & 0.00 & 46.17 & 57.16 \\ 
        + SWFC & 33.86 & 76.57 & 52.30 & 57.56 & 64.17 & 63.31 & 59.65 & 75.97 & 49,09 & 0.00 & 16.72 & 51.69 & 0.00 & 46.24 & 57.25 \\ 
        + SSLCL & 34.52 & 74.35 & 54.59 & 61.59 & 72.32 & 61.21 & 61.35 & 77.09 & 48.16 & 0.00 & 22.60 & 53.98 & 0.00 & 47.31 & 58.65 \\ 
        \midrule
        DialogueCRN & 31.79 & 66.84 & 58.31 & 63.72 & 72.17 & 60.45 & 60.87 & 77.04 & 48.40 & 0.00 & 24.32 & 51.50 & 0.00 & 47.16 & 58.39 \\ 
        + SupCon & 31.91 & 73.30 & 57.67 & 60.65 & 65.61 & 62.56 & 60.67 & 76.47 & 47.01 & 0.00 & 23.00 & 52.44 & 0.00 & 47.39 & 58.04 \\ 
        + mv-SupCon & 34.69 & 74.02 & 55.22 & 64.63 & 66.30 & 61.31 & 60.69 & 76.23 & 47.72 & 0.00 & 25.44 & 52.68 & 0.00 & 47.56 & 58.25 \\ 
        + SWFC & 34.87 & 73.27 & 55.36 & 64.78 & 68.35 & 61.83 & 61.14 & 76.08 & 48.14 & 0.00 & 27.95 & 53.07 & 0.00 & 46.19 & 58.30 \\ 
        + SSLCL & 36.60 & 74.82 & 58.72 & 65.68 & 70.09 & 61.98 & 62.78 & 76.82 & 55.50 & 0.00 & 28.97 & 53.23 & 0.00 & 49.04 & 59.93 \\ 
        \midrule
        HiTrans & 38.99 & 76.11 & 58.54 & 55.61 & 65.75 & 57.38 & 60.21 & 76.65 & 56.13 & 3.85 & 37.33 & 55.82 & 2.86 & 45.39 & 60.65 \\ 
        + SupCon & 36.55 & 68.90 & 59.56 & 52.94 & 71.02 & 54.32 & 59.12 & 76.61 & 54.91 & 0.00 & 34.22 & 55.31 & 0.00 & 48.80 & 60.47 \\ 
        + mv-SupCon & 39.43 & 72.21 & 54.99 & 55.41 & 73.68 & 52.24 & 59.05 & 77.04 & 53.95 & 0.00 & 37.91 & 51.61 & 0.00 & 42.30 & 59.44 \\ 
        + SWFC & 40.00 & 73.18 & 57.25 & 53.92 & 67.48 & 56.47 & 59.48 & 76.48 & 54.32 & 3.85 & 37.15 & 54.88 & 0.00 & 49.50 & 60.68 \\ 
        + SSLCL & 52.03 & 75.25 & 61.46 & 59.33 & 71.33 & 55.56 & 62.92 & 76.77 & 57.40 & 10.53 & 39.09 & 57.32 & 5.33 & 50.70 & 62.11 \\ 
        \midrule 
        $\text{COG-BART}^{\ast}$ & 27.76 & 80.89 & 64.49 & 60.82 & 63.92 & 66.67 & 63.75 & 79.44 & 57.78 & 12.90 & 38.20 & 61.50 & 2.90 & 49.59 & 63.82 \\ 
        + SupCon & 25.69 & 78.67 & 63.42 & 61.63 & 68.06 & 65.77 & 63.62 & 80.22 & 56.55 & 12.70 & 40.11 & 61.35 & 10.26 & 46.41 & 63.99 \\ 
        + SWFC & 28.90 & 79.67 & 62.99 &  60.33 & 69.32 & 64.68 & 63.79 & 79.09 & 57.27 & 18.46 & 41.23 & 61.08 & 18.42 & 46.15 & 63.85 \\ 
        + SSLCL & 31.25 & 79.95 & 66.92 & 62.46 & 71.90 & 65.59 & 65.88 & \textbf{80.52} & 59.40 & 18.35 & 41.55 & 63.61 & 27.45 & 46.67 & 65.49 \\ 
        \midrule
        DAG-ERC & 46.26 & 81.01 & 67.75 & 70.52 & 66.42 & 67.11 & 67.75 & 76.97 & 57.47 & 22.22 & 35.77 & 59.36 & 30.09 & 49.37 & 62.96 \\ 
        + SupCon & 47.39 & 80.33 & 67.48 & 69.86 & 67.90 & 67.72 & 68.03 & 77.07 & 57.76 & 21.24 & 36.55 & 59.63 & 32.26 & 49.51 & 63.20 \\ 
        + mv-SupCon & 47.02 & 80.74 & 67.43 & 71.39 & 65.11 & 66.85 & 67.49 & 76.76 & 57.76 & 27.20 & 36.31 & 60.20 & 29.27 & 48.54 & 63.02 \\ 
        + SWFC & 49.13 & 81.18 & 68.30 & 69.10 & 65.79 & 67.93 & 68.09 & 76.90 & 57.84 & 24.78 & 38.69 & 59.84 & 31.48 & 49.77 & 63.41 \\ 
        + SSLCL & 49.23 & 81.48 & \textbf{71.09} & 68.51 & 68.62 & 68.20 & 69.33 & 77.89 & 58.19 & 24.59 & 37.78 & 61.92 & \textbf{32.53} & 51.49 & 64.42 \\ 
        \midrule
        M3Net & 52.74 & 79.39 & 67.55 & 69.30 & 74.39 & 66.58 & 69.24 & 79.31 & 58.76 & 20.51 & 40.46 & 63.21 & 26.17 & 52.53 & 65.47 \\ 
        + SupCon & 48.80 & 80.17 & 66.67 & 67.68 & 75.62 & 66.58 & 68.86 & 78.58 & 59.50 & 23.19 & 39.56 & 65.04 & 23.16 & 52.70 & 65.40 \\ 
        + mv-SupCon & 51.23 & 80.26 & 66.17 & 69.01 & 69.40 & 67.25 & 68.12 & 78.11 & 59.72 & 23.08 & 42.05 & 63.60 & 23.53 & 53.91 & 65.34 \\ 
        + SWFC & 54.67 & 80.85 & 68.61 & 67.42 & 76.92 & 62.41 & 69.17 & 78.08 & 60.31 & 25.26 & 39.48 & 64.12 & 29.33 & 53.57 & 65.42 \\ 
        + SSLCL & \textbf{58.44} & \textbf{82.43} & 69.32 & \textbf{71.44} & \textbf{77.02} & \textbf{69.34} & \textbf{71.98} & 79.73 & \textbf{61.03} & \textbf{27.32} & \textbf{42.46} & \textbf{65.08} & 31.30 & \textbf{54.76} & \textbf{66.92} \\ 
        \midrule
        SPCL & 40.89 & 78.48 & 65.27 & 68.33 & 71.13 & 64.09 & 66.22 & 79.95 & 57.63 & 12.99 & 41.18 & 62.71 & 30.77 & 50.73 & 65.85 \\ 
        $\text{CKCL}^{\dagger}$ & - & - & - & - & - & - & 67.16 & - & - & - & - & - & - & - & 66.21 \\ 
        SCCL & 48.87 & 79.57 & 66.59 & 67.30 & 71.55 & 62.99 & 67.12 & 76.97 & 58.97 & 19.75 & 43.94 & 63.30 & 24.72 & 53.54 & 64.74 \\ 
      \bottomrule
  \end{tabular}}
  \caption{Experimental results on IEMOCAP and MELD. The best results in each emotion category and the overall w-F1 score are highlighted in bold. The $*$ symbol denotes that the original COG-BART implementation adopts mv-SupCon. The $\dagger$ symbol means that the source code of CKCL has not been released, and per-class w-F1 results are unavailable from the original paper.}
  \label{table:1}
\end{table*}

\subsection{Sample-Label Contrastive Learning}
\subsubsection{Soft-HGR as a Measure of Similarity} 
Soft-HGR aims to learn maximally correlated feature representations from different random variables. Motivated by the success of Soft-HGR in transfer learning \cite{8803726} and multimodal learning \cite{ma2021efficient, shi-huang-2023-multiemo}, we innovatively apply Soft-HGR to SCL as a similarity measure to effectively capture the  underlying correlations between sample features and label embeddings. Specifically, by regarding input samples and their corresponding emotion labels as two distinct random variables, we treat sample features and label embeddings as their respective feature mappings, which allows us to leverage Soft-HGR to learn sample features and label embeddings that are maximally correlated. Since a higher correlation indicates a greater similarity between feature mappings captured by Soft-HGR, we extend Soft-HGR as a measure of similarity between sample features and label embeddings.

Formally, give a batch of $N$ labeled training samples $\{(\textbf{u}_{i}^{t}, \textbf{u}_{i}^{a}, \textbf{u}_{i}^{v}, y_{i})\}^{N}_{i = 1}$, we denote the sample features learned through the ERC model as $\textbf{F} = [\textbf{f}_{1}, \textbf{f}_2, \dots, \textbf{f}_{N}]^{\mathrm{T}} \in \mathbb{R}^{N \times d}$, and label embeddings learned through the label embedding network as $\textbf{G} = [\textbf{g}_{1}, \textbf{g}_{2}, \dots, \textbf{g}_{K}]^{\mathrm{T}} \in \mathbb{R}^{K \times d}$. For an arbitrary $N$ discrete emotion labels $\textbf{z} = [z_{1}, z_{2}, \dots, z_{N}]^{\mathrm{T}} \in \mathbb{R}^{N}$, where $z_{i} \in \mathcal{Y}, 1 \leq i \leq N$, we denote their corresponding feature mappings as $\textbf{G}_{z} = [\textbf{g}_{z_{1}}, \textbf{g}_{z_{2}}, \dots, \textbf{g}_{z_{N}}]^{\mathrm{T}} \in \mathbb{R}^{N \times d}$. Based on the empirical Soft-HGR objective provided by \citeauthor{wang2019efficient} (2019), the Soft-HGR similarity between zero-meaned \footnote{This corresponds to the zero-mean constraint in Soft-HGR.} $\textbf{F}$ and $\textbf{G}_{z}$ can be calculated as follows:
\begin{equation}
\begin{aligned}
& \text{Sim}(\textbf{F}, \textbf{G}_{z}) = \frac{1}{N - 1}\sum_{i = 1}^{N}\textbf{f}_{i}^{\mathrm{T}}\textbf{g}_{z_{i}} - \frac{1}{2} \text{tr}(\text{cov}(\textbf{F})\text{cov}(\textbf{G}_{z})).
\end{aligned}  
\label{eq:5}
\end{equation}

Based on Equation \ref{eq:5}, the Soft-HGR similarity between a single sample-label pair $\textbf{f}_{i}$ and $\textbf{g}_{z_{i}}$ can be written as follows:
\begin{equation}
\begin{aligned}
& \text{Sim}(\textbf{f}_{i}, \textbf{g}_{z_{i}}) = \frac{1}{N - 1}\textbf{f}_{i}^{\mathrm{T}}\textbf{g}_{z_{i}} - \frac{1}{2}\sum_{l = 1}^{N}\text{cov}(\textbf{f}_{i}, \textbf{f}_{l})\text{cov}(\textbf{g}_{z_{i}}, \textbf{g}_{z_{l}}),
\end{aligned}  
\label{eq:6}
\end{equation}
where $\textbf{f}_{i} \in \textbf{F}, \textbf{g}_{z_{i}} \in \textbf{G}_{z}, 1 \leq i \leq N$.


\begin{table*}[t]
    \centering 
    \large  
    \resizebox{0.8\linewidth}{!}{
    \begin{tabular}{l|ccccccc}
      \toprule
      \multirow{2}{*}{Methods} & \multicolumn{7}{c}{IEMOCAP} \\ 
      & Happy & Sad & Neutral & Angry & Excited & Frustrated & \textbf{w-F1} \\
      \midrule
         SSLCL-based M3Net & 58.44 & 82.43 & 69.32 & 71.44 & 77.02 & 69.34 & 71.98 \\ 
        - Positive pairs augmentation & 54.48 & 81.13 & 67.55 & 66.67 & 75.90 & 68.22 & 70.04 \\
        - Negative sample-label loss & 55.56 & 80.18 & 67.36 & 68.79 & 76.53 & 68.41 & 70.33 \\ 
        - Label-label discrimination & 53.76 & 80.69 & 69.13 & 67.44 & 76.49 & 66.93 & 70.17 \\ 
      \bottomrule
  \end{tabular}}
  \caption{Ablation study of SSLCL on IEMOCAP to evaluate the impacts of positive pairs augmentation, negative sample-label loss, and label-label discrimination on model performances.}
  \label{table:3}
\end{table*}

\begin{table}[t]
    \centering 
    \large  
    \resizebox{1\linewidth}{!}{
    \begin{tabular}{c|c|cc}
      \toprule
      \multirow{2}{*}{Method} & \multirow{2}{*}{Similarity Measure} & IEMOCAP & MELD \\ 
      & & \textbf{w-F1} & \textbf{w-F1} \\
      \midrule
         \multirow{3}{*}{SSLCL-based M3Net} & Soft-HGR & 71.98 & 66.92 \\ 
         & Dot Product & 69.48 & 65.56 \\ 
         & Cosine Similarity & 69.70 & 65.61 \\ 
      \bottomrule
  \end{tabular}}
  \caption{Ablation study of SSLCL on IEMOCAP and MELD to evaluate the impact of different similarity measures on model performances.}
  \label{table:2}
\end{table}

\subsubsection{Positive Sample-Label Pairs} A pair $(\textbf{f}_{i}, \textbf{g}_{z_{i}})$ is referred to as a positive sample-label pair if $z_{i}$ is the ground-truth label of $\textbf{f}_{i}$. SSLCL aims to maximize the correlation between positive sample-label pairs. However, only one positive sample-label pair is available for each training sample, which could potentially hinder the generalizability and robustness of learned features. To address this limitation, inspired by Cutout \cite{devries2017improved}, a widely-used image augmentation technique in the computer vision community which masks out partial regions of input images during training, we leverage the multimodal nature of ERC utterances to generate additional positive sample-label pairs. Specifically, we consider utterances with one or two missing modalities as partially masked examples of the original input utterances, and utilize their corresponding learned sample features as augmentations of the original sample feature. Considering the crucial role of the textual modality in emotion classification, with audio and visual modalities serving as complementary cues \cite{shi-huang-2023-multiemo}, we strategically select the following missing modality scenarios as augmentations of $\textbf{f}_{i}$ : (1) $\textbf{f}_{i}(t)$: both the audio and visual modalities are missing; (2) $\textbf{f}_{i}(t + a)$: the visual modality is missing; (3) $\textbf{f}_{i}(t + v)$: the audio modality is missing. Importantly, our data augmentation technique does not alter the multimodal fusion mechanism employed by the ERC model.

Although the majority of existing ERC approaches make use of multimodal information, it is worth noting that several ERC methods are either bimodal-based \cite{10096885} or solely text-based \cite{li-etal-2020-hitrans, kim2021emoberta, li2022contrast}. In the case of bimodal ERC approaches, data augmentation is limited to the textual modality. Conversely, for text-based ERC models, data augmentation is not performed.

The augmented positive sample-label loss for the $i$-th sample can be calculated as follows:
\begin{align}
& \text{Aug}^{(i)} = \sum_{\tilde{\textbf{f}} \in F_{\text{aug}}^{(i)}} \exp (\text{Sim}(\tilde{\textbf{f}}, \textbf{g}_{z_{i}})), \\
& p_{\text{pos}}^{(i)} = \frac{\exp (\text{Sim}(\textbf{f}, \textbf{g}_{z_{i}}))}{\sum_{\textbf{g}_j \in \textbf{G}} \exp (\text{Sim}(\textbf{f}_{i}, \textbf{g}_j)) + \text{Aug}^{(i)}}, \\
& l_{\text{pos}}^{(i)} = - \sum_{\textbf{f} \in F_{\text{aug}}^{(i)} \cup \{\textbf{f}_{i}\}} \log (p_{\text{pos}}^{(i)}) (1 - p_{\text{pos}}^{(i)})^{\alpha},
\end{align}  
where $\alpha$ is a positive focusing parameter that forces the model to focus on hard positive examples, $F_{\text{aug}}^{(i)}$ is the set of augmented sample features for $\textbf{f}_{i}$. In the multimodal case, $F_{\text{aug}}^{(i)} = \{\textbf{f}_{i}(t), \textbf{f}_{i}(t + a), \textbf{f}_{i}(t + v)\}$. For the bimodal scenario, $F_{\text{aug}}^{(i)} = \{\textbf{f}_{i}(t)\}$. The text-based unimodal scenario has an empty augmentation set.

\begin{table*}[htbp]
    \centering 
    \large  
    \resizebox{1\linewidth}{!}{
    \begin{tabular}{c|c|ccccccc|cccccccc}
      \toprule
      \multirow{2}{*}{Method} & \multirow{2}{*}{LE Models} & \multicolumn{7}{c|}{IEMOCAP} & \multicolumn{8}{c}{MELD} \\ 
      &  & Happy & Sad & Neutral & Angry & Excited & Frustrated & \textbf{w-F1} & Neutral & Surprise & Fear & Sad & Joy & Disgust & Angry & \textbf{w-F1} \\
      \midrule
         \multirow{4}{*}{SSLCL-based M3Net} & Emb Layer & 55.56 & 80.18 & 67.36 & 68.79 & 76.53 & 68.41 & 70.33 & 78.78 & 60.54 & 20.51 & 40.74 & 64.90 & 23.26 & 55.60 & 66.02 \\ 
         & Two-Layer MLP & 58.44 & 82.43 & 69.32 & 71.44 & 77.02 & 69.34 & 71.98 & 79.73 & 61.03 & 27.32 & 42.46 & 65.08 & 31.30 & 54.76 & 66.92 \\ 
         & Three-Layer MLP & 61.34 & 80.75 & 69.91 & 69.34 & 78.51 & 67.62 & 71.77 & 79.80 & 61.56 & 29.44 & 41.48 & 65.46 & 28.28 & 55.43 & 67.05 \\ 
         & Bi-GRU & 62.14 & 81.42 & 70.37 & 68.66 & 78.02 & 69.47 & 72.33 & 79.97 & 61.07 & 31.21 & 41.64 & 65.23 & 32.07 & 55.19 & 67.15 \\ 
      \bottomrule
  \end{tabular}}
  \caption{Ablation study of SSLCL on IEMOCAP and MELD to evaluate the impact of different label embedding (LE) models on model performances. The three-layer MLP comprises an embedding layer followed by two fully-connected layers, while the Bi-GRU is composed of an embedding layer and two bidirectional GRUs. The hidden layer dimensions for both the three-layer MLP and Bi-GRU are tuned on the validation set.}
  \label{table:4}
\end{table*}

\subsubsection{Negative Sample-Label Pairs} A pair $(\textbf{f}_{i}, \textbf{g}_{z_{i}})$ is referred to as a negative sample-label pair if $z_{i}$ is not the ground-truth label of $\textbf{f}_{i}$. Unlike existing SCL approaches that do not directly compute the loss from negative pairs, SSLCL explicitly minimizes the similarity between negative sample-label pairs, which leads to improved model performances (refer to the Results and Analysis section). The negative sample-label loss for the $i$-th sample can be calculated as follows:
\begin{align}
& p_{\text{neg}}^{(i)} = \frac{\exp (\text{Sim}(\textbf{f}_{i}, \textbf{g}_{z_{i}}))}{\sum_{\textbf{g}_j \in \textbf{G}} \exp (\text{Sim}(\textbf{f}_{i}, \textbf{g}_j))}, \\
& l_{\text{neg}}^{(i)} = - \sum_{\textbf{g}_{z_{i}} \in G_{\text{neg}}^{(i)}} \log ( 1- p_{\text{neg}}^{(i)}){p_{\text{neg}}^{(i)}}^{\beta},
\end{align}  
where $\beta$ is a negative focusing parameter that assigns more focus to hard negative samples, $G_{\text{neg}}^{(i)} = \{\textbf{g}_k \in \textbf{G}|k \neq y_{i}\}$ is the set of negative label embeddings for $\textbf{f}_{i}$.

\subsubsection{Label-Label Discrimination} In addition to computing the loss of sample-label pairs, we introduce an auxiliary loss to minimize the similarity between all pairs of non-identical label embeddings. Experiments demonstrate that explicitly encouraging distinct representations for different label embeddings leads to considerable improvements in model performances (refer to the Results and Analysis section). The auxiliary loss is defined as follows:
\begin{align}
& p_{\text{label}}^{(i, j)} = \frac{\exp (\textbf{g}_i \cdot \textbf{g}_j)}{\sum_{\textbf{g}_k \in G_{i}} \exp (\textbf{g}_{i} \cdot \textbf{g}_k) + 1}, \label{eq:12}\\
& L_{\text{Label}} = - \sum_{\textbf{g}_{i} \in \textbf{G}} \sum_{\textbf{g}_{j} \in G_i} \log ( 1- p_{\text{label}}^{(i, j)}),
\end{align}  
where the $\cdot$ symbol represents dot product, $G_{i} = \{\textbf{g}_{j}|j \neq i\}$ is the set of label embeddings other than $\textbf{g}_{i}$. As shown in Equation \ref{eq:12}, when calculating the similarity probability between $\textbf{g}_{i}$ and $\textbf{g}_{j} \in G_{i}$, we explicitly set the similarity between $\textbf{g}_{i}$ and itself to 0 to prevent the value from becoming disproportionately large compared to dot products with other label embeddings, thus avoiding the softmax function from encountering regions with extremely small gradients.

\subsubsection{SSLCL Loss} In summary, the SSLCL loss for a batch of $N$ samples can be calculated as follows:
\begin{align}
L_{\text{SSLCL}} = \sum_{i = 1}^{N}{(\frac{N}{n_{y_{i}}})}^{\gamma}(l_{\text{pos}}^{(i)} + l_{\text{neg}}^{(i)}) + \lambda L_{\text{Label}},
\end{align}  
where $n_{y_{i}}$ is the count of label $y_{i}$ within the batch, $\gamma$ is a sample-weight parameter that assigns higher weights to minority emotions, $\lambda$ is a trade-off hyperparameter between the sample-label loss and the label-label loss.

\subsubsection{Cross-Entropy Loss} Apart from the SSLCL loss, we also adopt a cross-entropy (CE) loss to minimize the difference between predicted probabilities and ground-truth labels. Let $\textbf{p}_{i}$ denote the probability distribution of emotion classes for $\textbf{f}_{i}$, the CE loss can be defined as follows: 
\begin{align}
L_{\text{CE}} = -\sum_{i = 1}^{N}\log\textbf{p}_{i}[y_{i}], 
\end{align}  

\subsubsection{Overall Training Objective} The overall training objective is a linear combination of the SSLCL loss and the CE loss, which is defined as follows:
\begin{align}
L_{\text{Train}} = L_{\text{SSLCL}} + \eta L_{\text{CE}},
\end{align}  
where $\eta$ is a trade-off hyperparameter between the SSLCL loss and the CE loss.

\section{Experimental Settings}

\subsection{Datasets}
\subsubsection{IEMOCAP} The IEMOCAP dataset \cite{busso2008iemocap} consists of around 12 hours of dyadic conversation videos, which are divided into 151 dialogues with a total of 7433 individual utterances. Each utterance in IEMOCAP is annotated with one of six emotion labels: happy, sad, neutral, angry, excited, and frustrated.

\subsubsection{MELD} The MELD dataset \cite{poria-etal-2019-meld} is a collection of multi-party conversations extracted from the TV series \textit{Friends}, which consists of 1433 dialogues and a total of 13708 individual utterances. Each utterance in MELD is annotated with one of seven emotion categories: angry, disgust, fear, joy, neutral, sad, and surprise.

\subsection{Baseline Methods}
To demonstrate the compatibility and superiority of SSLCL compared to existing SCL methods, we conduct a comprehensive comparison between SSLCL and existing SCL literature in ERC, which can be categorized into two groups: (1) \textbf{Compatible SCL methods}: SCL approaches that can be integrated with the majority of existing ERC models, including vanilla SupCon loss \cite{khosla2020supervised}, multiview-SupCon (mv-SupCon) loss \cite{li2022contrast}, and SWFC loss \cite{shi-huang-2023-multiemo}; (2) \textbf{Incompatible SCL methods}: SCL frameworks that are incompatible with most existing ERC models, including SPCL \cite{song-etal-2022-supervised}, CKCL \cite{tu-etal-2023-context}, and SCCL \cite{yang2023cluster}.

Specifically, the effectiveness of SSLCL and compatible SCL approaches are compared on the following categories of ERC models: (1) \textbf{Recurrence-based ERC methods}: DialogueRNN \cite{majumder2019dialoguernn} and DialogueCRN \cite{hu-etal-2021-dialoguecrn}; (2) \textbf{Transformer-based ERC methods}: HiTrans \cite{li-etal-2020-hitrans} and COG-BART \cite{li2022contrast}; (3) \textbf{Graph-based ERC methods}: DAG-ERC \cite{shen-etal-2021-directed} and M3Net \cite{chen2023multivariate}.

As for incompatible SCL methods, due to the difficulty of applying them to most existing ERC models, we simply compare SSLCL with their original implementations. 

\subsection{Implementation Details}
\subsubsection{Evaluation Metrics} Following \citeauthor{shi-huang-2023-multiemo} (2023), we use the weighted-average F1 (w-F1) score to evaluate model performances on both IEMOCAP and MELD. All reported results are averaged over $5$ random seeds. 
\subsubsection{Modality Settings} In order to demonstrate the effectiveness of SSLCL across different modality settings, we leverage textual, audio and visual modalities for DialogueRNN, DialogueCRN and M3Net. Conversely, for text-based ERC models, specifically HiTrans, COG-BART and DAG-ERC, we solely utilize the textual modality. 
\subsubsection{Hyperparameter Settings} 
(1) \textbf{SSLCL Hyperparameters:} SSLCL requires minimal hyperparameter tuning across different ERC models. Specifically, the positive and negative focusing parameters $\alpha$ and $\beta$ are empirically set to $2.0$ and $0.5$ respectively, the sample-weight parameter $\gamma$ is chosen among $\{0.5, 1.0, 1.5\}$, the label loss trade-off hyperparameter $\gamma$ is selected among $\{0.5, 1.0, 2.0\}$, the CE trade-off hyperparameter $\eta$ is set to $1.0$, the label embedding network is optimized using Adam \cite{kingma2014adam} with a learning rate of $10^{-5}$, and its hidden dimension is designed to be twice the sample feature dimension learned by the ERC model. (2) \textbf{Hyperparameter Settings of Compared SCL Baselines:} To ensure a fair comparison, we implement both compatible and incompatible SCL baselines using the optimal hyperparameter values specified in their original papers. (3) \textbf{Hyperparameter Settings of ERC models:} In order to achieve a fair comparison, when comparing SSLCL with compatible SCL baselines on a ERC model, the hyperparameter settings of the ERC model are identical for all methods being evaluated. Specifically, to demonstrate that SSLCL can achieve superior performances without the need for a large batch size, we set the batch size to 4 on IEMOCAP and 8 on MELD for recurrence-based and graph-based ERC models, while for transformer-based ERC models, the batch size is set to 2 on IEMOCAP and 4 on MELD. 
\begin{table*}[t]
    \centering 
    \large  
    \resizebox{1\linewidth}{!}{
    \begin{tabular}{c|c|cccccc|cccccc}
      \toprule
      \multirow{2}{*}{Method} & \multirow{2}{*}{Training Strategy} & \multicolumn{6}{c|}{IEMOCAP (\textbf{w-F1})} & \multicolumn{6}{c}{MELD (\textbf{w-F1})} \\ 
      & & BS = 1 & BS = 2 & BS = 4 & BS = 8 & BS = 16 & BS = 32 & BS = 1 & BS = 2 & BS = 4 & BS = 8 & BS = 16 & BS = 32 \\
      \midrule
         \multirow{4}{*}{M3Net} & SupCon & 66.18 & 67.32 & 68.86 & 70.02 & 71.25 & 71.13 & 61.96 & 62.88 & 64.47 & 65.40 & 66.15 & 66.12 \\ 
         & mv-SupCon & 65.78 & 67.03 & 68.12 & 68.99 & 69.15 & 70.21 & 61.95 & 62.57 & 63.39 & 65.34 & 66.27 & 66.09 \\ 
          & SWFC & 65.56 & 67.95 & 69.17 & 70.83 & 71.10 & 70.99 & 62.11 & 62.98 & 64.32 & 65.42 & 66.28 & 66.29 \\ 
         & SSLCL & 69.88 & 71.12 & 71.98 & 72.25 & 72.33 & 72.31 & 65.13 & 66.08 & 66.55 & 66.92 & 67.14 & 67.20 \\ 
      \bottomrule
  \end{tabular}}
  \caption{Ablation study of SSLCL on IEMOCAP and MELD to demonstrate the capability of SSLCL in achieving stable model performances across different batch sizes. BS represents batch size.}
  \label{table:5}
\end{table*}

\section{Results and Analysis}
\subsection{Comparison with Baseline Methods}
The comparisons between SSLCL and existing SCL approaches on IEMOCAP and MELD are shown in Table \ref{table:1}. Experimental results conclusively demonstrate that SSLCL consistently brings noteworthy improvements to the performances of ERC models across different categories, achieving new state-of-the-art results not only in the overall w-F1 score but also across all emotion categories. Specifically, the average relative improvement SSLCL brought to ERC models are $3.32\%$ on IEMOCAP and $2.48\%$ on MELD, with SSLCL-based M3Net surpassing previous state-of-the-art approaches on both datasets. In contrast, under the constraint of a small batch size, existing compatible SCL methods only yield marginal or even detrimental effects across different ERC approaches. Furthermore,  SSLCL-integrated M3Net outperforms incompatible SCL frameworks by a significant margin on both IEMOCAP and MELD.  

\subsection{Quantitative Analysis of SSLCL}
\subsubsection{Impact of Positive Pairs Augmentation} As illustrated in Table \ref{table:3}, experimental results clearly demonstrate that removing positive pairs augmentation from SSLCL leads to a considerable decline in model performances. Particularly, there is a more noticeable decrease observed in minority emotions compared to majority emotions, such as \textit{Happy} and \textit{Angry}. The inferior performances of SSLCL without data augmentation verifies the effectiveness of positive pairs augmentation in enhancing model performances, especially in underrepresented emotion categories.

\subsubsection{Impact of Negative Sample-Label Loss} As shown in Table \ref{table:3},  when SSLCL does not directly calculate the loss of negative sample-label pairs, there is a significant decrease in performance across all emotion categories on IEMOCAP. This highlights the effectiveness of explicitly minimizing the similarity between negative sample-label pairs.

\subsubsection{Impact of Label-Label Discrimination} Based on Table \ref{table:3}, it is evident that removing the label-label loss from SSLCL leads to a considerable performance degradation across all emotion categories on IEMOCAP, particularly in non-neutral emotions. The results emphasize the significance of explicitly learning distinct representations for different label embeddings. 

\subsubsection{Impact of Soft-HGR Similarity Measure} Table \ref{table:2} demonstrates that Soft-HGR consistently outperforms dot product similarity and cosine similarity on both IEMOCAP and MELD, which highlights the superiority of Soft-HGR in effectively capturing the underlying correlations between sample features and label embeddings. 

\subsubsection{Impact of Label Embedding Network} To study the impact of label embedding network on model performances, we implement SSLCL-based M3Net with different label embedding (LE) models, including a single embedding (Emb) layer, a two-layer MLP, a three-layer MLP (an embedding layer followed by two fully-connected layers) and Bi-GRU (an embedding layer followed by two bidirectional GRUs), on both IEMOCAP and MELD. As illustrated in Table \ref{table:4}, the performance of the single embedding layer is considerably poorer compared to other LE models, while the two-layer MLP achieves a comparable performance to the three-layer MLP. Notably, Bi-GRU outperforms the other compared methods on both IEMOCAP and MELD. Nevertheless, it is important to note that the improvement of Bi-GRU over the two-layer MLP is not substantial, especially considering the significantly higher computational cost associated with Bi-GRU compared to the two-layer MLP. Therefore, in order to trade off between model performance and efficiency, we select the two-layer MLP as the label embedding network in SSLCL.

\subsection{Batch Size Stability of SSLCL} To investigate the stability of SSLCL across different batch sizes, we conduct experiments on both MELD and IEMOCAP using the M3Net. As shown in Table \ref{table:5}, SSLCL consistently outperforms existing compatible SCL approaches by a significant margin across all batch size settings, with a more pronounced improvement in small batch sizes. The results clearly demonstrate the superiority of SSLCL over existing SCL methods in achieving remarkable model performances without being constrained by the batch size.

\section{Conclusion}
In this paper, we propose an efficient and model-agnostic SCL framework named SSLCL for the ERC task, in which a novel utilization of label representation is designed to address the challenges with large batch sizes and model incompatibility encountered in existing SCL approaches. Additionally, we innovatively leverage Soft-HGR as a measure of similarity, which significantly outperforms traditional similarity measures. Furthermore, multimodal information is innovatively utilized by SSLCL as data augmentation to boost model performances. Experimental results on IEMOCAP and MELD demonstrate the compatibility and superiority of SSLCL over existing state-of-the-art SCL approaches.

\bibliography{aaai24}

\begin{thebibliography}{33}
\providecommand{\natexlab}[1]{#1}

\bibitem[{Arumugam, Bhattacharjee, and Yuan(2022)}]{9938005}
Arumugam, B.; Bhattacharjee, S.~D.; and Yuan, J. 2022.
\newblock Multimodal Attentive Learning for Real-time Explainable Emotion Recognition in Conversations.
\newblock In \emph{2022 IEEE International Symposium on Circuits and Systems (ISCAS)}, 1210--1214.

\bibitem[{Bao et~al.(2019)Bao, Li, Huang, Zhang, Zheng, Zamir, and Guibas}]{8803726}
Bao, Y.; Li, Y.; Huang, S.-L.; Zhang, L.; Zheng, L.; Zamir, A.; and Guibas, L. 2019.
\newblock An Information-Theoretic Approach to Transferability in Task Transfer Learning.
\newblock In \emph{2019 IEEE International Conference on Image Processing (ICIP)}, 2309--2313.

\bibitem[{Busso et~al.(2008)Busso, Bulut, Lee, Kazemzadeh, Mower, Kim, Chang, Lee, and Narayanan}]{busso2008iemocap}
Busso, C.; Bulut, M.; Lee, C.-C.; Kazemzadeh, A.; Mower, E.; Kim, S.; Chang, J.~N.; Lee, S.; and Narayanan, S.~S. 2008.
\newblock IEMOCAP: Interactive emotional dyadic motion capture database.
\newblock \emph{Language resources and evaluation}, 42: 335--359.

\bibitem[{Chen et~al.(2023)Chen, Shao, Zhu, and Shen}]{chen2023multivariate}
Chen, F.; Shao, J.; Zhu, S.; and Shen, H.~T. 2023.
\newblock Multivariate, Multi-Frequency and Multimodal: Rethinking Graph Neural Networks for Emotion Recognition in Conversation.
\newblock In \emph{Proceedings of the IEEE/CVF Conference on Computer Vision and Pattern Recognition}, 10761--10770.

\bibitem[{DeVries and Taylor(2017)}]{devries2017improved}
DeVries, T.; and Taylor, G.~W. 2017.
\newblock Improved regularization of convolutional neural networks with cutout.
\newblock \emph{arXiv preprint arXiv:1708.04552}.

\bibitem[{Gebelein(1941)}]{gebelein1941statistische}
Gebelein, H. 1941.
\newblock Das statistische Problem der Korrelation als Variations-und Eigenwertproblem und sein Zusammenhang mit der Ausgleichsrechnung.
\newblock \emph{ZAMM-Journal of Applied Mathematics and Mechanics/Zeitschrift f{\"u}r Angewandte Mathematik und Mechanik}, 21(6): 364--379.

\bibitem[{Ghosal et~al.(2019)Ghosal, Majumder, Poria, Chhaya, and Gelbukh}]{ghosal-etal-2019-dialoguegcn}
Ghosal, D.; Majumder, N.; Poria, S.; Chhaya, N.; and Gelbukh, A. 2019.
\newblock {D}ialogue{GCN}: A Graph Convolutional Neural Network for Emotion Recognition in Conversation.
\newblock In \emph{Proceedings of the 2019 Conference on Empirical Methods in Natural Language Processing and the 9th International Joint Conference on Natural Language Processing (EMNLP-IJCNLP)}, 154--164. Hong Kong, China: Association for Computational Linguistics.

\bibitem[{Hirschfeld(1935)}]{hirschfeld1935connection}
Hirschfeld, H.~O. 1935.
\newblock A connection between correlation and contingency.
\newblock In \emph{Mathematical Proceedings of the Cambridge Philosophical Society}, volume~31, 520--524. Cambridge University Press.

\bibitem[{Hu, Wei, and Huai(2021)}]{hu-etal-2021-dialoguecrn}
Hu, D.; Wei, L.; and Huai, X. 2021.
\newblock {D}ialogue{CRN}: Contextual Reasoning Networks for Emotion Recognition in Conversations.
\newblock In \emph{Proceedings of the 59th Annual Meeting of the Association for Computational Linguistics and the 11th International Joint Conference on Natural Language Processing (Volume 1: Long Papers)}, 7042--7052. Online: Association for Computational Linguistics.

\bibitem[{Hu et~al.(2021)Hu, Liu, Zhao, and Jin}]{hu-etal-2021-mmgcn}
Hu, J.; Liu, Y.; Zhao, J.; and Jin, Q. 2021.
\newblock {MMGCN}: Multimodal Fusion via Deep Graph Convolution Network for Emotion Recognition in Conversation.
\newblock In \emph{Proceedings of the 59th Annual Meeting of the Association for Computational Linguistics and the 11th International Joint Conference on Natural Language Processing (Volume 1: Long Papers)}, 5666--5675. Online: Association for Computational Linguistics.

\bibitem[{Khosla et~al.(2020)Khosla, Teterwak, Wang, Sarna, Tian, Isola, Maschinot, Liu, and Krishnan}]{khosla2020supervised}
Khosla, P.; Teterwak, P.; Wang, C.; Sarna, A.; Tian, Y.; Isola, P.; Maschinot, A.; Liu, C.; and Krishnan, D. 2020.
\newblock Supervised contrastive learning.
\newblock \emph{Advances in neural information processing systems}, 33: 18661--18673.

\bibitem[{Kim and Vossen(2021)}]{kim2021emoberta}
Kim, T.; and Vossen, P. 2021.
\newblock EmoBERTa: Speaker-Aware Emotion Recognition in Conversation with RoBERTa.
\newblock arXiv:2108.12009.

\bibitem[{Kingma and Ba(2014)}]{kingma2014adam}
Kingma, D.~P.; and Ba, J. 2014.
\newblock Adam: A method for stochastic optimization.
\newblock \emph{arXiv preprint arXiv:1412.6980}.

\bibitem[{Lee and Lee(2022)}]{lee-lee-2022-compm}
Lee, J.; and Lee, W. 2022.
\newblock {C}o{MPM}: Context Modeling with Speaker{'}s Pre-trained Memory Tracking for Emotion Recognition in Conversation.
\newblock In \emph{Proceedings of the 2022 Conference of the North American Chapter of the Association for Computational Linguistics: Human Language Technologies}, 5669--5679. Seattle, United States: Association for Computational Linguistics.

\bibitem[{Li et~al.(2020)Li, Ji, Li, Zhang, and Liu}]{li-etal-2020-hitrans}
Li, J.; Ji, D.; Li, F.; Zhang, M.; and Liu, Y. 2020.
\newblock {H}i{T}rans: A Transformer-Based Context- and Speaker-Sensitive Model for Emotion Detection in Conversations.
\newblock In \emph{Proceedings of the 28th International Conference on Computational Linguistics}, 4190--4200. Barcelona, Spain (Online): International Committee on Computational Linguistics.

\bibitem[{Li, Yan, and Qiu(2022)}]{li2022contrast}
Li, S.; Yan, H.; and Qiu, X. 2022.
\newblock Contrast and generation make bart a good dialogue emotion recognizer.
\newblock In \emph{Proceedings of the AAAI conference on artificial intelligence}, volume~36, 11002--11010.

\bibitem[{Luo, Phan, and Reiss(2023)}]{10096885}
Luo, J.; Phan, H.; and Reiss, J. 2023.
\newblock Cross-Modal Fusion Techniques for Utterance-Level Emotion Recognition from Text and Speech.
\newblock In \emph{ICASSP 2023 - 2023 IEEE International Conference on Acoustics, Speech and Signal Processing (ICASSP)}, 1--5.

\bibitem[{Ma, Huang, and Zhang(2021)}]{ma2021efficient}
Ma, F.; Huang, S.-L.; and Zhang, L. 2021.
\newblock An efficient approach for audio-visual emotion recognition with missing labels and missing modalities.
\newblock In \emph{2021 IEEE international conference on multimedia and Expo (ICME)}, 1--6. IEEE.

\bibitem[{Majumder et~al.(2019)Majumder, Poria, Hazarika, Mihalcea, Gelbukh, and Cambria}]{majumder2019dialoguernn}
Majumder, N.; Poria, S.; Hazarika, D.; Mihalcea, R.; Gelbukh, A.; and Cambria, E. 2019.
\newblock Dialoguernn: An attentive rnn for emotion detection in conversations.
\newblock In \emph{Proceedings of the AAAI conference on artificial intelligence}, volume~33, 6818--6825.

\bibitem[{Poria et~al.(2017)Poria, Cambria, Hazarika, Majumder, Zadeh, and Morency}]{poria-etal-2017-context}
Poria, S.; Cambria, E.; Hazarika, D.; Majumder, N.; Zadeh, A.; and Morency, L.-P. 2017.
\newblock Context-Dependent Sentiment Analysis in User-Generated Videos.
\newblock In \emph{Proceedings of the 55th Annual Meeting of the Association for Computational Linguistics (Volume 1: Long Papers)}, 873--883. Vancouver, Canada: Association for Computational Linguistics.

\bibitem[{Poria et~al.(2019)Poria, Hazarika, Majumder, Naik, Cambria, and Mihalcea}]{poria-etal-2019-meld}
Poria, S.; Hazarika, D.; Majumder, N.; Naik, G.; Cambria, E.; and Mihalcea, R. 2019.
\newblock {MELD}: A Multimodal Multi-Party Dataset for Emotion Recognition in Conversations.
\newblock In \emph{Proceedings of the 57th Annual Meeting of the Association for Computational Linguistics}, 527--536. Florence, Italy: Association for Computational Linguistics.

\bibitem[{R{\'e}nyi(1959)}]{renyi1959measures}
R{\'e}nyi, A. 1959.
\newblock On measures of dependence.
\newblock \emph{Acta mathematica hungarica}, 10(3-4): 441--451.

\bibitem[{Shen et~al.(2021)Shen, Wu, Yang, and Quan}]{shen-etal-2021-directed}
Shen, W.; Wu, S.; Yang, Y.; and Quan, X. 2021.
\newblock Directed Acyclic Graph Network for Conversational Emotion Recognition.
\newblock In \emph{Proceedings of the 59th Annual Meeting of the Association for Computational Linguistics and the 11th International Joint Conference on Natural Language Processing (Volume 1: Long Papers)}, 1551--1560. Online: Association for Computational Linguistics.

\bibitem[{Shi and Huang(2023)}]{shi-huang-2023-multiemo}
Shi, T.; and Huang, S.-L. 2023.
\newblock {M}ulti{EMO}: An Attention-Based Correlation-Aware Multimodal Fusion Framework for Emotion Recognition in Conversations.
\newblock In \emph{Proceedings of the 61st Annual Meeting of the Association for Computational Linguistics (Volume 1: Long Papers)}, 14752--14766. Toronto, Canada: Association for Computational Linguistics.

\bibitem[{Song et~al.(2022)Song, Huang, Xue, and Hu}]{song-etal-2022-supervised}
Song, X.; Huang, L.; Xue, H.; and Hu, S. 2022.
\newblock Supervised Prototypical Contrastive Learning for Emotion Recognition in Conversation.
\newblock In \emph{Proceedings of the 2022 Conference on Empirical Methods in Natural Language Processing}, 5197--5206. Abu Dhabi, United Arab Emirates: Association for Computational Linguistics.

\bibitem[{Sun et~al.(2017)Sun, Wei, Ren, and Ma}]{sun2017label}
Sun, X.; Wei, B.; Ren, X.; and Ma, S. 2017.
\newblock Label Embedding Network: Learning Label Representation for Soft Training of Deep Networks.
\newblock arXiv:1710.10393.

\bibitem[{Tu et~al.(2023)Tu, Liang, Mao, Yang, and Xu}]{tu-etal-2023-context}
Tu, G.; Liang, B.; Mao, R.; Yang, M.; and Xu, R. 2023.
\newblock Context or Knowledge is Not Always Necessary: A Contrastive Learning Framework for Emotion Recognition in Conversations.
\newblock In \emph{Findings of the Association for Computational Linguistics: ACL 2023}, 14054--14067. Toronto, Canada: Association for Computational Linguistics.

\bibitem[{Tu et~al.(2022)Tu, Wen, Liu, Jiang, and Cambria}]{9706271}
Tu, G.; Wen, J.; Liu, C.; Jiang, D.; and Cambria, E. 2022.
\newblock Context- and Sentiment-Aware Networks for Emotion Recognition in Conversation.
\newblock \emph{IEEE Transactions on Artificial Intelligence}, 3(5): 699--708.

\bibitem[{Wang et~al.(2019)Wang, Wu, Huang, Zheng, Xu, Zhang, and Huang}]{wang2019efficient}
Wang, L.; Wu, J.; Huang, S.-L.; Zheng, L.; Xu, X.; Zhang, L.; and Huang, J. 2019.
\newblock An efficient approach to informative feature extraction from multimodal data.
\newblock In \emph{Proceedings of the AAAI Conference on Artificial Intelligence}, volume~33, 5281--5288.

\bibitem[{Wu et~al.(2018)Wu, Xiong, Yu, and Lin}]{wu2018unsupervised}
Wu, Z.; Xiong, Y.; Yu, S.~X.; and Lin, D. 2018.
\newblock Unsupervised feature learning via non-parametric instance discrimination.
\newblock In \emph{Proceedings of the IEEE conference on computer vision and pattern recognition}, 3733--3742.

\bibitem[{Yang et~al.(2023{\natexlab{a}})Yang, Gao, Wu, Gan, Ding, Jiang, and Nie}]{yang-etal-2023-self}
Yang, H.; Gao, X.; Wu, J.; Gan, T.; Ding, N.; Jiang, F.; and Nie, L. 2023{\natexlab{a}}.
\newblock Self-adaptive Context and Modal-interaction Modeling For Multimodal Emotion Recognition.
\newblock In \emph{Findings of the Association for Computational Linguistics: ACL 2023}, 6267--6281. Toronto, Canada: Association for Computational Linguistics.

\bibitem[{Yang et~al.(2023{\natexlab{b}})Yang, Zhang, Alhuzali, and Ananiadou}]{yang2023cluster}
Yang, K.; Zhang, T.; Alhuzali, H.; and Ananiadou, S. 2023{\natexlab{b}}.
\newblock Cluster-level contrastive learning for emotion recognition in conversations.
\newblock \emph{IEEE Transactions on Affective Computing}.

\bibitem[{Zhang, Chen, and Chen(2023)}]{zhang-etal-2023-dualgats}
Zhang, D.; Chen, F.; and Chen, X. 2023.
\newblock {D}ual{GAT}s: Dual Graph Attention Networks for Emotion Recognition in Conversations.
\newblock In \emph{Proceedings of the 61st Annual Meeting of the Association for Computational Linguistics (Volume 1: Long Papers)}, 7395--7408. Toronto, Canada: Association for Computational Linguistics.

\end{thebibliography}

\end{document}